\documentclass[conference]{IEEEtran}
\usepackage{cite}
\usepackage{graphicx}
\usepackage{multirow}
\usepackage{url}
\usepackage[caption=false,font=footnotesize]{subfig}
\usepackage{booktabs}
\usepackage{array}
\newcolumntype{C}[1]{>{\centering\arraybackslash}p{#1}} % centered p-column
\usepackage{amsthm}
\usepackage[cmex10]{amsmath}
\usepackage{comment}
\usepackage{amsfonts}
\usepackage{amssymb}
\graphicspath{{figures/}}
\usepackage{bm}
\usepackage{epstopdf}
\usepackage{float}
\usepackage{xcolor}

\usepackage[T1]{fontenc}
\newcommand*{\affmark}[1][*]{\textsuperscript{\dag}}
\usepackage{algorithm}
\usepackage[noend]{algpseudocode}
\makeatletter
\def\BState{\State\hskip-\ALG@thistlm}
\makeatother
\usepackage{acronym}
\acrodef{ISAC}[ISAC]{Integrated Sensing and Communication}
\definecolor{revisionred}{RGB}{180,0,0}
\colorlet{red}{black}
\colorlet{revisionred}{black}
\colorlet{violet}{black}
\definecolor{trackingred}{RGB}{0,0,0}

\newcommand{\panelcaption}[1]{\par\vspace{0.2em}{\centering\footnotesize\textcolor{black}{#1}\par}}

\ifCLASSINFOpdf
  \else  
\fi

\usepackage[normalem]{ulem}

\begin{document}
\title{\Large{\textcolor{trackingred}{5G NR Sensing of Robotic Arm Motion in Indoor Factory Environments}}}

\author{\IEEEauthorblockN{Neeraj Varshney$^{1,2}$, Anuraag Bodi$^{1,2}$, Steve Blandino$^{1,2}$, Jian Wang$^{1}$, Camillo Gentile$^{1}$, and Nada Golmie$^{1}$}
\IEEEauthorblockA{$^{1}$National Institute of Standards and Technology
Gaithersburg, Maryland USA\\$^{2}$Prometheus Computing LLC, Bethesda, MD}}

\maketitle

\begin{abstract}
This paper investigates the performance of monostatic Integrated Sensing and Communication (ISAC) systems for tracking robotic arm kinematics in industrial factory (InF) environments. Building on a part-specific quasi-deterministic channel model previously derived from high-resolution robotic-arm measurements using our context-aware channel sounder and digital twin, we develop a framework for generating arbitrary articulated motion profiles through hierarchical kinematic transformations. By integrating this high-fidelity target model with the standardized 3GPP InF channel model, we employ 5G New Radio (NR) Positioning Reference Signals (PRS) as the sensing waveform to enable comprehensive system-level evaluation. Simulations are conducted for both sparse clutters (InF-SH) and dense clutters (InF-DH) scenarios using 3GPP-compliant antenna radiation patterns and analog beamforming, allowing us to assess the impact of industrial clutter on sensing performance. The results show that while high estimation accuracy is achievable under Line-of-Sight (LOS) conditions, dense clutter in InF-DH environments introduces severe blockage that significantly limits sensing reliability. In particular, the effective sensing coverage in InF-DH is drastically reduced, with detection probability dropping to approximately \(P_d \approx 10\,\%\) even when the robotic arm is located directly beneath the Transmission Reception Point (TRP). Furthermore, in InF-SH, achieving a high detection probability (e.g., \(P_d \geq 90\,\%\)) requires the target to remain in close proximity to the TRP, with an effective coverage radius of approximately 13 m.
\end{abstract}


% \fi
\IEEEpeerreviewmaketitle
\vspace{-0.1cm}
\section{Introduction}
\vspace{-0.1cm}

The integration of sensing capabilities into 5th-generation New Radio (5G NR) and future networks represents a transformative shift, positioning cellular infrastructure as a multi-functional foundation for a wide range of industrial and commercial applications~\cite{qaisar2026role}. Originally designed for high-throughput and low-latency communications, the inherent waveform structure, beamforming capabilities, and high-resolution reference signals of 5G NR make it uniquely suitable for integrated sensing and communication (ISAC) tasks~\cite{3GPP38765,varshney2026isac}. In industrial environments, the ability to monitor high-precision machinery, such as robotic arms, without the need for dedicated, costly radar hardware is critical for enhancing situational awareness and operational safety~\cite{3GPP22837,zandi2023robot}. By leveraging standardized 5G infrastructure, industrial facilities can achieve scalable, cost-effective monitoring solutions that complement existing automation frameworks~\cite{5GAmericasISAC}.

The primary motivation for this study stems from the need to reliably detect and track articulated targets in complex Indoor Factory (InF) environments, which are characterized by dense clutter and challenging line-of-sight (LOS) conditions. Prior works in \cite{blandino2025detecting,blandino2026evaluation,sahoo2025cooperative} have investigated 5G-based sensing for drone detection in 3GPP Urban Macro/Micro with Aerial Vehicles (UMa-AV/UMi-AV) scenarios. However, a fundamental understanding of the detection and positioning limits for multi-segment articulated targets, such as robotic arms, in standardized 3GPP InF scenarios remains lacking. Standardized 3GPP channel models, such as 3GPP TR 38.901~\cite{3GPP38901}, provide a robust framework for simulating these industrial scenarios, but they often lack the part-specific detail required to capture the dynamic scattering behavior of moving mechanical joints. Recent studies have demonstrated the feasibility of using Positioning Reference Signals (PRS) for target localization in outdoor scenarios~\cite{blandino2025detecting,blandino2026evaluation,sahoo2025cooperative,fernndez2026integrated,varshney2026multi}, yet the interplay between industrial clutter density, elevated Transmission Reception Point (TRP) deployment, and articulated target kinematics in InF scenarios requires more comprehensive evaluation.

{\color{red}The primary contribution of this paper is a system-level evaluation of 5G NR PRS-based monostatic sensing for detecting the motion of an articulated robotic arm in standardized 3GPP InF environments. \textcolor{trackingred}{To enable this evaluation, we build on our measurement-derived, part-specific Quasi-Deterministic (QD) robotic-arm target model~\cite{bodi2025digital}. The model was developed from high-resolution radio frequency (RF), camera, and Light Detection and Ranging (LiDAR) measurements that construct a digital twin and associate measured multipath components with individual articulating parts, yielding experimentally grounded part-specific scattering statistics. Here, we extend the model with hierarchical kinematic transformations to generate arbitrary, mechanically consistent motion trajectories~\cite{nist_qdrarm}.} We then combine the resulting time-varying target response with the standardized 3GPP InF background channel and a 3GPP-compliant antenna and beamforming configuration. This combination is not presented as a replacement for, or an extension of, the 3GPP InF channel model: the 3GPP model describes the factory background propagation, whereas the QD model supplies the articulated target response required for sensing evaluation. Using this simulation methodology, we evaluate time-resolved Doppler signatures, spatial detection probability, and conditional velocity and localization errors in sparse and dense clutter environments. The results quantify how clutter blockage, target distance, and directional antenna gain constrain the sensing coverage of a monostatic TRP.}

%The remainder of this paper is organized as follows: Section \ref{Sec:SystemModel} introduces the 5G NR monostatic radar system model, incorporating the 3GPP Release-19 InF ISAC channel framework. Section \ref{Sec:targetchannel} details the proposed robotic arm channel simulator, including the kinematic transformation model and the part-specific quasi-deterministic channel formulation. Section \ref{Sec:Evaluation} presents the evaluation results, spatial coverage analysis, and performance insights across sparse and dense industrial environments. Finally, Section \ref{Sec:Conclusion} concludes the paper.
\vspace{-0.14cm}
\section{System Model}
\vspace{-0.14cm}
\label{Sec:SystemModel}
{\color{red}In this work, we consider a monostatic ISAC system designed to sense the motion of a robotic arm in an industrial setting. The deployment follows the 3GPP NR ISAC InF scenario~\cite{3GPP38901}, in which the network comprises $N_{\mathrm{TRP}}=18$ TRPs arranged in a square lattice. Because multi-TRP fusion is outside the scope of this work, the sensing performance is evaluated using one representative TRP, denoted as TRP~$i$, from the 18-TRP deployment. The selected TRP is equipped with $N$ co-located transmit and receive antenna elements arranged in a uniform rectangular array (URA) to enable analog beamforming in both azimuth and elevation. It operates in monostatic sensing mode using a cyclic prefix-orthogonal frequency division multiplexing (CP-OFDM) waveform and employs downlink PRS for delay and Doppler estimation, target-motion detection, and localization. In this study, the PRS resources are used exclusively for sensing evaluation, and communication traffic is not considered.}
\vspace{-0.15cm}
\subsection{Transmit Signal Model with Analog Beamforming}
\vspace{-0.05cm}
The TRP operates as an OFDM radar where PRS sequences are mapped onto the resource grid following a comb-type mapping with spacing~$K$. After OFDM modulation and cyclic prefix insertion, a frequency-flat analog beamforming vector~$\mathbf{f}_i \in \mathbb{C}^{N\times1}$ is applied to steer the sensing energy toward the target region. The transmitted signal from TRP~$i$ on subcarrier~$k$ and OFDM symbol~$\ell$ is
%\begin{equation}
$\mathbf{x}_{i,k,\ell} = \mathbf{f}_i s_{k,\ell}\in \mathbb{C}^{N\times1},$
%\end{equation}
where $s_{k,\ell}$ is the known PRS symbol. The beamforming vector is normalized such that $\|\mathbf{f}_i\|^2 = P_{\mathrm{TX}}$ to maintain constant symbol transmit power. 
%Sensing observations are collected over a coherent processing interval (CPI) of $T_{\mathrm{CPI}} = N_{\mathrm{CPI}} T_{\mathrm{PRS}}$.
\vspace{-0.1cm}
\subsection{ISAC Channel Model for Robotic Arm Sensing}
\vspace{-0.1cm}
The propagation channel at TRP $i$ is modeled as a superposition of reflections from the articulated sensing target (ST) i.e., the robotic arm and the monostatic background (BG) clutter  
\vspace{-0.1cm}
\begin{equation}
\mathbf{H}_i(\tau,t) = \mathbf{H}^{(\mathrm{ST})}_i(\tau,t) + \mathbf{H}^{(\mathrm{BG})}_i(\tau,t),
\end{equation}
where $\tau$ and $t$ denote delay and time, respectively. {\color{red}The standardized 3GPP InF model characterizes factory background propagation, blockage, and clutter, but does not provide the time-varying, part-specific scattering response of an articulated robotic arm; the QD target model supplies this missing response~\cite{bodi2025digital}. Therefore, we superpose the separately generated target and background responses for sensing evaluation without modifying or extending the standardized 3GPP InF channel model.}
The robotic arm is modeled as a multi-segment sensing target comprising four rigid articulating segments (base, shoulder, upper arm, and forearm). Each segment is represented by its geometric centroid derived from its bounding box. The target channel for robotic arm is formulated in Sec.~\ref{Sec:targetchannel}. 

%Mechanical fault detection is facilitated by monitoring anomalies in the time-varying Doppler phase shifts and centroid trajectories $\mathbf{x}_q(t)$.

The background channel $\textbf{H}^{(\mathrm{BG})}(\tau,t)$ characterizes reflections from static environmental clutter, such as industrial machinery and storage racks. Following the 3GPP monostatic sensing framework, this is modeled as the coherent sum of responses from $N_{\mathrm{RP}}=3$ virtual reference points (RPs)
\begin{equation}
{\color{trackingred}
\textbf{H}^{(\mathrm{BG})}(\tau,t)
=
\sum_{p=1}^{N_{\mathrm{RP}}}
10^{-(PL_{i,p}+SF_{i,p})/20}
\textbf{H}^{(\mathrm{RP})}_{i,p}(\tau,t), \label{bgChannel}
}
\end{equation}
{\color{trackingred}where $PL_{i,p}$ and $SF_{i,p}$ represent path loss and shadow fading (both in dB) associated with the $p$th RP with respect to TRP~$i$.} For a stationary TRP and static clutter, the background channel remains time-invariant over the coherent processing interval (CPI). To accurately represent the InF environment, the 3D coordinates of these RPs are determined by drawing the 2D distance and height from gamma distributions respectively, $\Gamma(\alpha_d, \beta_d) + c_d$ and $\Gamma(\alpha_h, \beta_h) + c_h$, using parameters specifically calibrated for the InF scenario~\cite{3GPP38901}.

\vspace{-0.1cm}
\subsection{Received Signal and Doppler Processing}
\vspace{-0.1cm}
At the receiver, analog combining is applied using the vector~$\mathbf{w}_i \in \mathbb{C}^{N\times1}$. Leveraging the monostatic configuration where transmit and receive paths are identical, the received symbol after OFDM demodulation is
%\begin{equation}
$\mathbf{y}_{i,k,\ell} =
\mathbf{w}_{i}^{H}
\left(
\mathbf{H}_{i,k,\ell}\mathbf{f}_i s_{k,\ell}
+
\mathbf{n}_{i,k,\ell}
+
\boldsymbol{\xi}_{i,k,\ell}
\right)$,
%\end{equation}
where $\mathbf{n}_{i,k,\ell}$ represents thermal noise, and $\boldsymbol{\xi}_{i,k,\ell}$ denotes residual self-interference. For sensing processing, the received PRS symbols are collected over a CPI. Each PRS occasion contains $L_{\mathrm{PRS}}$ PRS OFDM symbols and repeats every $T_{\mathrm{PRS}}$ seconds. A CPI consists of $N_{\mathrm{CPI}}$ consecutive PRS occasions, yielding a total CPI duration of $T_{\mathrm{CPI}} = N_{\mathrm{CPI}} T_{\mathrm{PRS}}$. The sensing receiver at TRP $i$ processes the PRS resource elements to estimate the effective channel response. A least-squares (LS) estimate of the frequency-domain channel is obtained via matched filtering with the known PRS symbols as
% \begin{equation}
$\hat{\mathbf{g}}_{i,k,\ell} =
\frac{1}{\sqrt{P_{\mathrm{TX}}}}
\frac{\mathbf{y}_{i,k,\ell} s_{k,\ell}^{*}}
{|s_{k,\ell}|^{2} + \epsilon},$
% \end{equation}
where $s_{k,\ell}$ denotes the transmitted PRS symbol and $\epsilon$ is a small regularization constant to ensure numerical stability. The estimated channel responses across the PRS subcarriers are transformed into complex range profiles through an inverse FFT of size $N_R$, producing $r_i[n,m]$, where $n$ denotes the range bin and $m$ indexes the PRS occasions within a coherent processing interval (CPI). To suppress static background clutter, slow-time mean subtraction is applied
\begin{equation}
\tilde{r}_i[n,m] =
r_i[n,m] -
\frac{1}{N_{\mathrm{CPI}}}
\sum_{m'=0}^{N_{\mathrm{CPI}}-1}
r_i[n,m'] .
\end{equation}

Doppler processing is then performed by applying an $N_D$-point FFT across the PRS-occasion index $m$ at each range bin using a Doppler window $w_D[m]$. This produces the complex range–Doppler map per RF chain
\begin{equation}
X_{\textcolor{trackingred}{q}}[n,\nu] =
\frac{1}{\sqrt{N_D}}
\sum_{m=0}^{N_{\mathrm{CPI}}-1}
w_D[m]\,
\tilde{r}_{\textcolor{trackingred}{q}}[n,m]\,
e^{-j2\pi m\nu/N_D}.
\end{equation}
Stacking the outputs across the $N_{\mathrm{RF}}$ receiver chains forms the range–Doppler snapshot vector
%\begin{equation}
$\mathbf{X}[n,\nu] =
\left[
X_1[n,\nu],\ldots,X_{N_{\mathrm{RF}}}[n,\nu]
\right]^T $.
%\end{equation}
The Doppler frequency associated with bin $\nu$ is given by
%\begin{equation}
$\hat{f}_D[\nu] =
\frac{\nu}{N_D T_{\mathrm{PRS}}}$,
%\end{equation}
which corresponds to the estimated radial velocity
%\begin{equation}
$\hat{v}_r[\nu] =
\frac{\lambda}{2}\hat{f}_D[\nu]
=
\frac{\lambda \nu}{2 N_D T_{\mathrm{PRS}}}$.
%\end{equation}
It is important to note that the Doppler signatures of the articulated arm segments reflect their instantaneous velocities. During normal operation, the motion of the robotic segments produces non-zero Doppler components. However, when a mechanical fault occurs at time $t_f$ and the motion ceases, the radial velocity approaches zero, causing the Doppler energy to concentrate near the zero-Doppler bin. Tracking this abrupt shift of Doppler energy over a period of time therefore provides a reliable indicator of motion seizure and enables sensing-based fault detection.

\vspace{-0.1cm}
\section{Robotic Arm Channel Simulator}
\label{Sec:targetchannel}
To model the sensing characteristics of articulated targets in industrial environments, we adopt a two-step approach. First, a kinematic model is developed to capture the time-varying geometry of the robotic arm, followed by the incorporation of this model into a quasi-deterministic channel framework.
\vspace{-0.1cm}
\subsection{Deterministic centroids based on Kinematic Model}
\begin{figure*}[t!]
\centering
\subfloat[\textcolor{red}{Hierarchical joint rotations, including the reference geometry and sequential transformations with $\boldsymbol{\Theta} = [90^\circ, 90^\circ, 90^\circ]$.}\label{fig:kinematicmodel}]{%
    \includegraphics[width=0.55\textwidth]{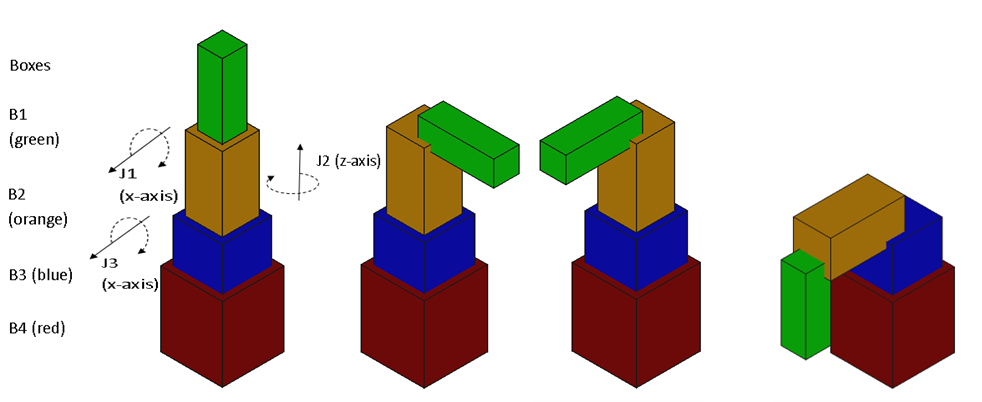}}
    \hspace{0.5cm}
\subfloat[\textcolor{red}{Simulated trajectory over time.}\label{fig:RoboticArmMotion}]{%
    \includegraphics[width=0.23\textwidth]{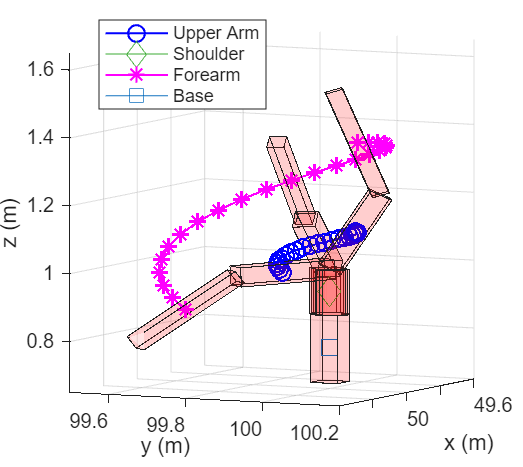}}
    \vspace{-0.2cm}
\caption{\textcolor{red}{Robotic-arm kinematic model and motion generation: (a) hierarchical rigid-body transformations and (b) an example articulated trajectory.}}
\label{fig:ArmKinematicsAndMotion}
\vspace{-0.5cm}
\end{figure*}

{\color{red}The arm comprises four rigid cuboids, $B_1$--$B_4$, representing the forearm, upper arm, shoulder, and stationary base, respectively, as illustrated in Fig.~\ref{fig:kinematicmodel}. The base origin $\mathbf o=[x_0,y_0,z_0]^T$ places the complete target in the global coordinate system, while the eight vertices of each cuboid describe its local geometry and spatial extent. The arm pose is specified by $\boldsymbol{\Theta}=[J_1,J_2,J_3]$, with $J_1$ rotating the forearm about the $x$-axis and $J_2$ and $J_3$ rotating the upstream links about the $z$-axis and $x$-axis, respectively. The joint limits are $J_2\in[-180^\circ,180^\circ]$ and $J_1,J_3\in[-90^\circ,90^\circ]$. Each homogeneous vertex $\widetilde{\mathbf v}=[x,y,z,1]^T$ is transformed as $\widetilde{\mathbf v}'=\mathbf M(\boldsymbol{\Theta})\widetilde{\mathbf v}$, where $\mathbf M$ combines the appropriate rotation and translation matrices. The products are applied hierarchically: a proximal-joint rotation is inherited by all distal links, whereas a distal-joint rotation affects only its associated link. This dependency preserves link connectivity and produces mechanically consistent articulated poses rather than four independently moving point targets.}

{\color{red}At time $t_k$, the transformed vertices of segment $l$ are reduced to the centroid
$\mathbf c_l(t_k)=\frac{1}{8}\sum_{i=1}^{8}\mathbf v_{l,i}(t_k)$, $l\in\{1,2,3,4\}$. These four time-varying centroids locate the part-specific scattering regions used by the channel model. A motion realization is generated by drawing feasible initial and final joint angles and linearly interpolating them over the observation interval. The transformation chain and centroids are recomputed at every time step, providing position and velocity trajectories for each link; Fig.~\ref{fig:RoboticArmMotion} shows one realization. The kinematic component is therefore deterministic once the motion profile is chosen, whereas the channel component described in Section~\ref{QDCHannel} introduces stochastic scattering around the resulting segment locations. This separation permits repeatable motion experiments while retaining realistic propagation variability.}
\vspace{-0.1cm}
\subsection{Quasi-deterministic Sensing Target Channel}
\label{QDCHannel}
\begin{table}[t!]
\caption{\textcolor{red}{Path-loss and shadow-fading parameters~\cite{3GPP38901}.}}
\label{tab:InF_PL_params}
\centering
\resizebox{\columnwidth}{!}{%
\begin{tabular}{@{}llcc@{}}
\toprule
\textbf{Condition}  & \textbf{Path Loss $PL(d_{3D},f_c)$ [dB]} & $\boldsymbol{\sigma_{SF}}$ \textbf{[dB]} \\ \midrule
\textbf{LOS} All InF & $31.84+21.50\log_{10}(d_{3D})+19.00\log_{10}(f_c)$ & 4.3 \\ \midrule
\textbf{NLOS}  InF-SH & $\max(PL_{\mathrm{LOS}},32.4+23.0\log_{10}(d_{3D})+20\log_{10}(f_c))$ & 5.9 \\
 \textbf{NLOS} InF-DH & $\max(PL_{\mathrm{LOS}},33.63+21.9\log_{10}(d_{3D})+20\log_{10}(f_c))$ & 4.0 \\ \bottomrule
\end{tabular}%
}
\parbox{\columnwidth}{ Note: Carrier frequency $f_c$ is in GHz and 3D distance $d_{3D}$ is in meters ($1\leq d_{3D}\leq600$~m).}
\vspace{-0.7cm}
\end{table}

To characterize the robotic-arm sensing response, we adopt the measurement-derived, part-specific QD channel model in~\cite{bodi2025digital} where each articulating segment $l$ is represented by the geometric centroid $\mathbf{c}_{l}(t)$ of its corresponding bounding box. The total target-induced impulse response is formulated as the superposition of the time-varying contributions from the four rigid segments\textcolor{trackingred}{,} the base, shoulder, upper arm, and forearm\textcolor{trackingred}{,} and is given by the following expression
\begin{align}
    H^{(\mathrm{ST})}(\tau,t) = &\sum_{l=1}^{4} \sum_{k=1}^{N_{l}(t)} \alpha_{lk}(t) \mathbf{a}_{rx}(\Omega_{lk}^{AOA}) \nonumber\\&\times\mathbf{a}_{tx}^{H}(\Omega_{lk}^{AOD}) \delta(\tau - \tau_{lk}).
\end{align}
Here, $N_{l}(t)$ represents the number of stochastic paths generated around each centroid according to a Poisson distribution. For each path $k$, the scattering point $\mathbf{c}_{lk}(t)$ is defined by adding a stochastic 3D displacement $\Delta \mathbf{c}_{lk}(t)=r_{lk}(t)\cdot\mathbf{u}_{lk}(t)$ to the segment centroid, where $r_{lk}(t)$ is the Gamma-distributed radial offset and $\mathbf{u}_{lk}(t)$ is a random direction vector drawn uniformly over unit sphere. The absolute delay $\tau_{lk}$ and the angles of arrival ($\Omega_{lk}^{AOA}$) and departure ($\Omega_{lk}^{AOD}$) are derived from the instantaneous 3D Cartesian coordinates of these scattering points relative to the transceiver positions. The complex path amplitude is computed as $\alpha_{lk}(t) = G_{lk}(t)e^{j\phi_{lk}(t)}$, where the path gain $G_{lk}$ in dB, incorporates the two-way path loss $PL_{dB}$, shadow fading $\sigma_{SF}$, and the time-varying Radar Cross Section (RCS) $\sigma_{lk}(t)$ scaled by the number of paths:
\begin{align}
    G_{lk}^{dB}(t) = -\bigg[ &2(PL_{dB} + \sigma_{SF}) - \left(\sigma_{lk}(t) - 10\log_{10}N_l(t)\right) \nonumber\\&+ 10\log_{10}\left(\frac{\lambda^2}{4\pi}\right) \bigg]. \label{eq:G_lk}
\end{align}
%$G_{lk}^{dB}(t) = -[ 2(PL_{dB} + \sigma_{SF}) - (\sigma_{lk}(t) - 10\log_{10}N_l(t)) + 10\log_{10}(\frac{\lambda^2}{4\pi})].$
Here, $PL_{\mathrm{dB}}$ is modeled as a function of the carrier frequency $f_c$ and the 3D distance $d_{3D}$ between the TRP and the scattering point. The specific path loss models and shadow fading standard deviations ($\sigma_{SF}$) for the 3GPP InF scenarios, specifically InF-SH and InF-DH, are summarized in Table \ref{tab:InF_PL_params} \cite{3GPP38901}. Furthermore, the associated phase $\phi_{lk}(t)$ of the complex path amplitude incorporates a random initial phase $\phi_0$ and a distance-dependent factor $\phi_{lk}(t) = \left[ \phi_0 + \frac{2\pi}{\lambda}(d_{lk}^{TX}(t) + d_{lk}^{RX}(t)) \right] \pmod{2\pi}$, where $\lambda$ denotes the wavelength and $d_{lk}^{TX}(t), d_{lk}^{RX}(t)$ represent the time-varying distances from the transmitter and receiver to the target centroid, respectively. This modeling approach ensures that the deterministic component strictly adheres to the robot arm's mechanical motion limits, while the stochastic variations accurately capture the measured scattering characteristics of the individual articulated parts. 

The LOS condition in Table \ref{tab:InF_PL_params} for these elevated TRP configurations is determined by a distance-dependent exponential function. The probability of a LOS condition existing over a 2D distance $d_{2D}$ is expressed as $\text{Pr}_{\mathrm{LOS}} = \exp(-d_{2D} / k_{subsce})$. This probability is governed by a scaling factor $k_{subsce}$, which incorporates physical environment parameters including the typical clutter size $d_{clutter}$, clutter density $r$ which is the percentage of floor area covered by structural blockages such as machinery, and the effective clutter height $h_c$. Specifically, for the InF-SH and InF-DH sub-scenarios where the target scattering point height $h_{\mathrm{ST}}$ is below the clutter height $h_c$, the scaling factor is defined as
%\begin{equation}
$k_{subsce} = -\frac{d_{clutter}}{\ln(1-r)} \cdot \frac{h_{\mathrm{TRP}} - h_{\mathrm{ST}}}{h_c - h_{\mathrm{ST}}}$.
%\end{equation}
It should be noted that if the target is positioned at or above the effective clutter height ($h_{\mathrm{ST}} \ge h_c$), the LOS probability is set to unity ($\text{Pr}_{\mathrm{LOS}} = 1$), reflecting a permanent line-of-sight condition between the elevated TRP and the target as per the 3GPP industrial channel model specifications.

\section{Evaluation Results and Insights}
\label{Sec:Evaluation}
In this section, we present system-level evaluation results obtained using our open-source NR ISAC simulation framework~\cite{nist_5gnrad}. The evaluation is conducted under the 3GPP InF scenario, incorporating a robotic-arm channel modeling framework to capture realistic sensing dynamics~\cite{nist_qdrarm}. To evaluate the proposed ISAC framework across diverse industrial conditions, we consider two representative 3GPP InF scenarios as reported in Table \ref{tab:InF_HighBS_params}. The first configuration, InF-SH, models a large factory hall ($300~\text{m} {\times} 150~\text{m} {\times} 10$~m) characterized by sparse clutter. The second configuration, InF-DH, represents a smaller ($120~\text{m}  {\times} 60~\text{m} {\times} 10$~m) environment with dense clutter. Although the full network comprises $N_{\mathrm{TRP}}=18$ TRPs in a square lattice arrangement, our comprehensive sensing performance analysis focuses on a single TRP. This TRP is located at the Global Coordinate System (GCS) positions of $[125, 75]$~m in the InF-SH scenario and $[50, 30]$~m in the InF-DH scenario. This specific TRP serves as a representative reference to characterize the monostatic sensing link's behavior under varying industrial clutter conditions.

\begin{table}[t!]
\caption{Parameters for Indoor Factory High BS Scenarios \cite{3GPP38901}}
\vspace{-0.35cm}
\label{tab:InF_HighBS_params}
\centering
\begin{tabular}{@{}lcc@{}}
\toprule
\textbf{Parameter} & \textbf{InF-SH (Sparse)} & \textbf{InF-DH (Dense)} \\ \midrule
BS height $h_{BS}$ (m) & 8 & 8 \\
ISD (m) & 50 & 20 \\
Clutter size $d_{clutter}$ (m) & 10 & 2 \\
Clutter density $r$ & 0.20 (20~\%) & 0.60 (60~\%) \\
Effective clutter height $h_c$ (m) & 2 & 6 \\ \midrule
\end{tabular}
\vspace{-0.8cm}
\end{table}

The selected TRP operates in a monostatic configuration where the sensing transmitter and sensing receiver are co-located. It is modeled as an elevated base station with an antenna height of $h_{\mathrm{TRP}} = 8$~m, placing it above the dominant clutter layers ($h_{\mathrm{TRP}} > h_c$) to maximize sensing coverage. The system operates at a carrier frequency of $f_c = 4$~GHz with a bandwidth of $B = 100$~MHz and a subcarrier spacing of $\Delta f = 30$~kHz. Sensing is performed using downlink NR PRS signals on a CP-OFDM waveform, configured with a comb size of $K = 2$ and $L_{\mathrm{PRS}} = 2$ symbols per occasion. Unless otherwise stated, Doppler estimation is performed over a CPI of $N_{\mathrm{CPI}} = 128$ occasions, yielding a total observation duration of $T_{\mathrm{CPI}} = 128$~ms to accurately track robotic arm kinematics. The TRP is equipped with an $8\times8$ URA comprising $\pm45^\circ$ dual-polarized antenna elements. The inter-element spacing is set to $(d_H, d_V) = (0.5\lambda, 0.5\lambda)$ along the horizontal and vertical dimensions, respectively. The element radiation pattern follows the 3GPP TR~38.901 model~\cite{3GPP38901}, with a maximum element gain of $8$~dBi and a $65^\circ$ half-power beamwidth in both azimuth and elevation. Analog beamforming is applied across the array to steer directional beams toward the target. 
\begin{figure}[t!]
\centering
\includegraphics[height=1.3in,width=2.4in]{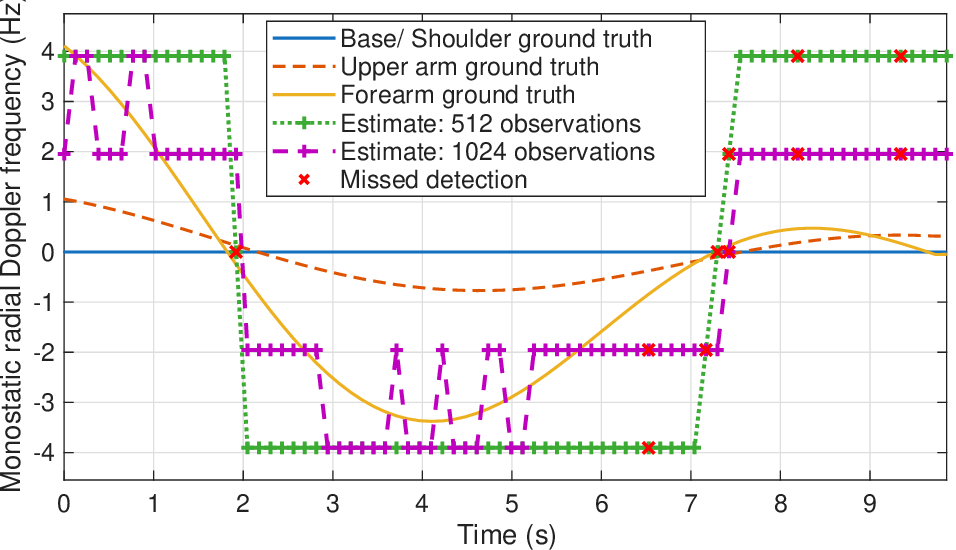}
\vspace{-0.4cm}
\caption{\textcolor{red}{Time-resolved ground-truth and estimated monostatic radial Doppler frequencies of the robotic arm. Here, the non-zero Doppler frequencies corresponding to missed detections are interpolated for better understanding.}}
\label{fig:DopplerVsTime}
\vspace{-.7cm}
\end{figure}

The robotic arm is modeled as a physical sensing target situated approximately 1~m above the ground. Because the arm height remains below the effective clutter height ($h_{\mathrm{ST}} < h_c$) in both scenarios, the arm remains embedded within the clutter region throughout its motion. This placement creates a challenging sensing scenario, where the direct path may be partially obstructed and multipath components play a significant role in detection and tracking. To derive comprehensive performance insights across the coverage area, we define a circular evaluation grid\footnote{In the large-scale InF-SH scenario, locations are generated within a circular region of radius $65$~m using a spatial resolution of $2$~m. For the dense InF-DH environment, the sensing region is reduced to a $30$~m radius with a finer resolution of $1$~m to capture higher spatial variability.} centered at the TRP, where each grid point represents an independent simulation drop. For each simulation drop, the arm follows a randomized motion trajectory for 10~seconds to evaluate dynamic performance. The trajectory is sampled every 128~ms, resulting in 78 temporal states per location. At each time step, the 3D position and velocity of the arm are updated, inducing time-varying propagation distances and Doppler shifts that are captured by the ISAC system.

Fig.~\ref{fig:DopplerVsTime} presents the time-resolved monostatic radial Doppler response of the robotic arm over the 10~s motion interval, with the arm state updated every 128~ms. The ground-truth Doppler frequency of each articulated component is obtained as $f_{\mathrm{D}}=2f_c v_r/c$, where $v_r$ is its radial velocity relative to the sensing TRP. The base/shoulder remains stationary at 0~Hz, whereas the upper arm and forearm exhibit distinct time-varying Doppler trajectories; the forearm reaches approximately $\pm4$~Hz because of its larger radial velocity. At each sensing instant, the estimated Doppler is selected from the highest-power detected peak in the range--Doppler map. Processing with 512 and 1024 coherent slow-time observations, spaced by 0.5~ms, corresponds to CPI durations of 256 and 512~ms and nominal Doppler resolutions of 3.91 and 1.95~Hz, respectively. Thus, the 1024-observation configuration better represents intermediate Doppler states at the cost of a longer coherent observation interval. Because the dominant reflection can originate from different arm segments, the strongest estimate need not continuously follow one component but instead represents the dominant instantaneous scatterer. Red crosses denote missed detections, which occur when no target-associated peak satisfies the detection criterion because of unfavorable orientation, fading, multipath interference, static-clutter masking, or Doppler spreading. Despite these intermittent missed detections i.e., 7 and 4 missed out of 78  for 512 and 1024 slow-time observations respectively, the time-varying motion of the robotic arm is successfully detected over most of the observation interval.

\begin{figure}[t!]
\centering
\begin{minipage}[t]{0.24\textwidth}
    \centering
    \includegraphics[width=\linewidth]{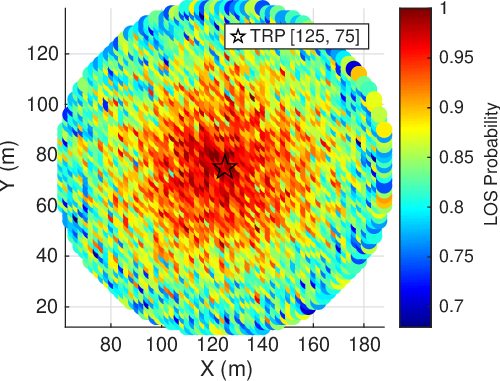}
    \panelcaption{(a) InF-SH}
\end{minipage}
\begin{minipage}[t]{0.24\textwidth}
    \centering
    \includegraphics[width=\linewidth]{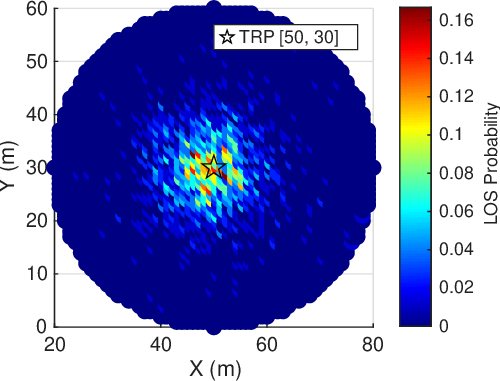}
    \panelcaption{(b) InF-DH}
\end{minipage}
\vspace{-0.2cm}
\caption{Impact of environment on LOS probability distribution.}
\label{fig:losProb}
\vspace{-0.5cm}
\end{figure}

The impact of the environment is clearly reflected in the spatial distribution of the LOS probability, as shown in Fig.~\ref{fig:losProb}. In the InF-SH scenario, the combination of sparse clutter and larger obstacle dimensions results in a consistently high LOS probability across most of the sensing region, thereby enabling reliable sensing. In this regime, the overall sensing performance is primarily limited by the intrinsic system capabilities rather than environmental factors. In contrast, the InF-DH scenario exhibits a rapid degradation in LOS probability due to dense obstructions, which significantly attenuate the direct sensing path. Consequently, sensing performance in such dense environments becomes very challenging.

Next, the spatial coverage and detection probability ($P_d$) for robotic arm motion are evaluated across both sparse and dense industrial environments. To rigorously characterize these results, we define two key performance metrics. First, the \textit{expected coverage} is defined as the ideal sensing range extending up to half of the inter-site distance (ISD), i.e.,
% \begin{equation}
    $R_{\mathrm{exp}} = \frac{D}{2}$,
% \end{equation}
which represents the theoretical boundary within which a TRP is expected to maintain reliable sensing before the target transitions to the coverage region of a neighboring cell. Second, the \textit{effective sensing coverage} is defined as the maximum distance from the TRP at which the detection probability remains higher than $P_{d,\text{Required}}$, i.e.,
\begin{equation}
    R_{\mathrm{eff}} = \max \left\{ r \geq 0 \mid P_d(r) > P_{d,\text{Required}} \right\},
\end{equation}
where $r$ denotes the 2D radial distance between the TRP and the sensing target. This metric captures the practical sensing range by accounting for channel impairments such as path loss, shadow fading, and blockage effects.
\begin{figure}[t!]
\centering
\begin{minipage}[t]{0.22\textwidth}
    \centering
    \includegraphics[width=\linewidth]{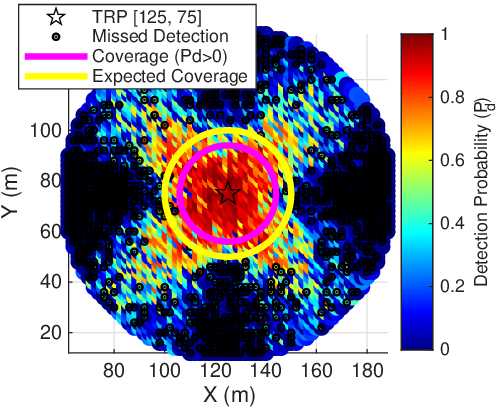}
    \panelcaption{(a) InF-SH}
\end{minipage}
\begin{minipage}[t]{0.22\textwidth}
    \centering
    \includegraphics[width=\linewidth]{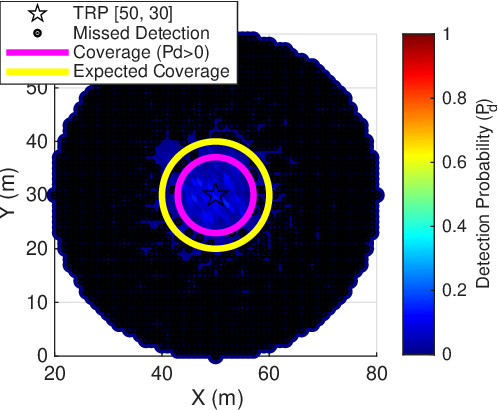}
    \panelcaption{(b) InF-DH}
\end{minipage}
\caption{Detection performance under different environments.}
\vspace{-0.5cm}
\label{fig:detPer}
\end{figure}
\begin{figure}[t!]
\centering
\includegraphics[width=0.5\linewidth]{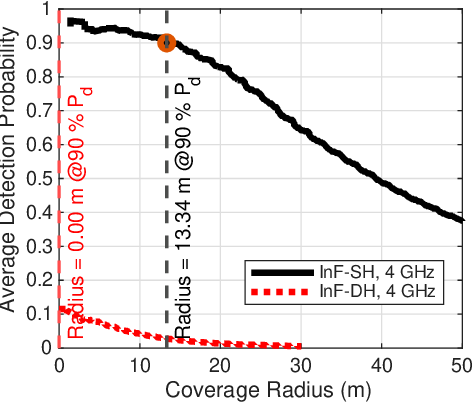}
\vspace{-0.2cm}
\caption{Average detection performance w.r.t. coverage radius.}
\label{fig:AvgdetPer}
\vspace{-0.65cm}
\end{figure}

Fig.~\ref{fig:detPer}(a-b) illustrates the detection performance under the InF-SH and InF-DH scenarios, where $P_d$ is defined as $P_d = \frac{TP}{TP + FN}$, with $TP$ and $FN$ denoting the number of true positive and false negative detections, respectively. First, it is evident from Fig.~\ref{fig:detPer} that the sparse clutter and larger obstacle dimensions in the InF-SH scenario enable relatively high $P_d$ values compared to the InF-DH scenario, where dense clutter and increased effective clutter height introduce severe blockage effects. Second, $P_d$ attains its maximum when the robotic arm is located directly beneath the TRP and gradually decreases as the distance from the TRP increases. This trend is particularly prominent in the InF-SH scenario, where the LOS conditions are more favorable.
\begin{figure*}[t!]
\centering
\begin{minipage}[t]{0.24\textwidth}
    \centering
    \includegraphics[width=\linewidth]{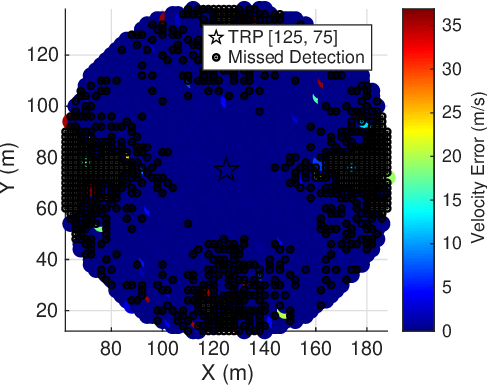}
    \panelcaption{(a) InF-SH velocity}
\end{minipage}
\begin{minipage}[t]{0.24\textwidth}
    \centering
    \includegraphics[width=\linewidth]{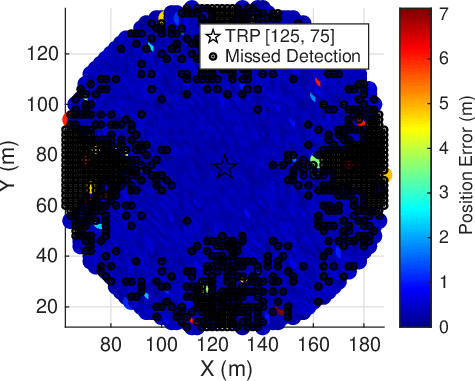}
    \panelcaption{(b) InF-SH position}
\end{minipage}
\begin{minipage}[t]{0.24\textwidth}
    \centering
    \includegraphics[width=\linewidth]{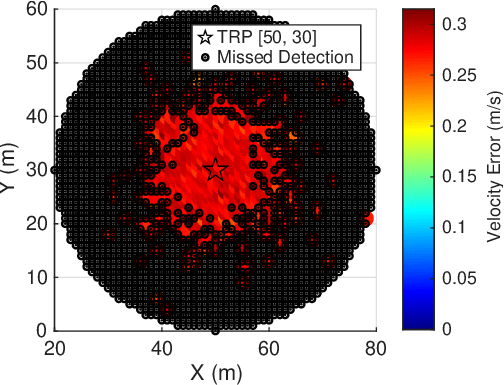}
    \panelcaption{(c) InF-DH velocity}
\end{minipage}
\begin{minipage}[t]{0.24\textwidth}
    \centering
    \includegraphics[width=\linewidth]{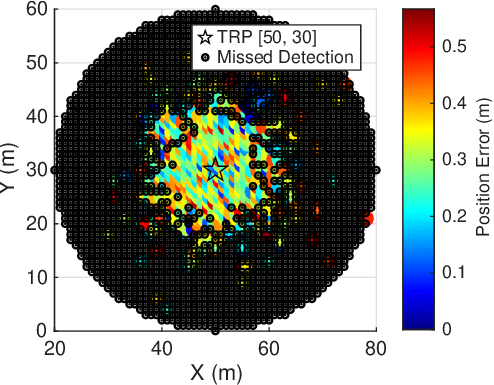}
    \panelcaption{(d) InF-DH position}
\end{minipage}
\caption{\textcolor{red}{Velocity and localization errors in the InF-SH and InF-DH environments.}}
\label{fig:Error}
\vspace{-0.65cm}
\end{figure*}
Interestingly, a distinct ``butterfly'' pattern can be observed in the detection probability distribution for the InF-SH scenario. This phenomenon arises due to missed detections of the complete arm motion at certain spatial locations. The underlying cause is primarily attributed to the $8 \times 8$~URA antenna configuration combined with the analog beamforming architecture. Specifically, the use of dual-polarized elements along with fixed phase weights to achieve vertical electrical tilt results in a non-isotropic radiation pattern. Consequently, the antenna gain is concentrated in specific angular lobes, leading to enhanced detection sensitivity along certain azimuth directions, while reduced gain or nulls appear in others. As a result of these directional gain variations and associated missed detections, a noticeable performance gap as shown in Fig.~\ref{fig:detPer}(a) emerges between the expected and effective sensing coverage, even when considering a relaxed threshold of $P_{d,\text{Required}} = 0~\%$. When a stricter requirement of $P_{d,\text{Required}} = 90~\%$ is imposed, the effective sensing coverage in the InF-SH scenario further reduces to $13.34$~m, as shown in Fig.~\ref{fig:AvgdetPer}. In contrast, the InF-DH scenario fails to meet this performance requirement altogether due to the compounded effects of severe blockage and limited LOS availability. Overall, this gap highlights the inherent limitations of monostatic sensing in complex industrial environments, where both environmental factors and antenna characteristics jointly constrain the achievable sensing performance.

Fig.~\ref{fig:Error} presents the velocity and 3D localization error performance across the InF-SH and InF-DH scenarios. Each point in the heatmap corresponds to an aggregated error metric computed over the full motion trajectory of the robotic arm at a given spatial location. Specifically, the velocity error is evaluated as the root mean square error (RMSE) between the ground truth and estimated velocity, while the position error is computed as the RMSE of the 3D localization over time. It is important to note that these metrics are calculated only for realizations in which the target is successfully detected, thereby reflecting the conditional estimation accuracy of the system. In the InF-SH scenario, the results show relatively low and spatially uniform errors, particularly within the effective coverage around the TRP, owing to favorable LOS conditions and stronger received signal power. However, as the distance from the TRP increases, localized increases in both velocity and position errors are occasionally observed, primarily due to reduced SNR and enhanced multipath-induced estimation uncertainty. In contrast, the InF-DH scenario exhibits highly concentrated regions of low error near the TRP, while most of the sensing area is dominated by missed detections. This indicates that, although accurate estimation is achievable in localized regions with sufficient signal quality, the overall sensing reliability is severely constrained by dense clutter and blockage. These results highlight an important insight: while estimation accuracy remains high when detections occur, the dominant performance bottleneck in dense industrial environments is not estimation error but detection availability due to LOS blockage. This reinforces the need for improved spatial diversity, advanced beamforming strategies, or multi-TRP cooperation to ensure consistent sensing performance across the coverage area.

\section{Conclusion}
\label{Sec:Conclusion}
{\color{red}We evaluated 5G NR PRS-based monostatic sensing of an articulated robotic arm using a part-specific QD target model and standardized 3GPP InF background channels. The time-resolved Doppler result distinguishes the stationary base from moving links, while the coverage study shows that sparse InF-SH supports substantially more reliable sensing than dense InF-DH. When detection occurs, estimation errors remain low; the dominant limitation is sensing-link availability under blockage and directional antenna gain. Improved beamforming, spatial diversity, and multi-TRP cooperation are therefore important for reliable industrial sensing. The simulator can also support future data-driven fault-detection studies~\cite{nist_qdrarm}.}

\bibliographystyle{IEEEtran}
\bibliography{references.bib}

\end{document}